\documentclass[10pt,twocolumn,letterpaper]{article}

\usepackage[pagenumbers]{main}

\usepackage[table]{xcolor}
\usepackage{microtype}
\usepackage{graphicx}
\usepackage{booktabs}
\usepackage{amsmath}
\usepackage{amssymb}
\usepackage{mathtools}
\usepackage{amsthm}
\usepackage{pifont}

\usepackage{color}
\usepackage{enumitem}
\usepackage{fontawesome5}
\usepackage{multirow}
\usepackage{makecell}

\usepackage{algorithmic}
\usepackage{algorithm}
\usepackage{etoolbox,siunitx}

\definecolor{lblue}{RGB}{231, 66, 52}

\usepackage{csquotes}
\usepackage{bm}

\usepackage{pgfplots}
\pgfplotsset{width=10cm,compat=1.9}
\usepackage{tikz}
\usetikzlibrary{
    positioning,
    calc,
    decorations.pathreplacing,
    shapes.geometric
}

\usepackage[pagebackref,breaklinks,colorlinks,citecolor=lblue]{hyperref}

\def\paperID{1}
\def\confName{RoboSense}
\def\confYear{IROS 2025}

\newcommand{\bestresult}[1]{\text{\textbf{#1}}}

\title{GBlobs: Local LiDAR Geometry for Improved Sensor Placement Generalization}

\author{
    Du\v{s}an Mali\'c$^{1,2}$ \qquad
    Christian Fruhwirth-Reisinger$^{1,2}$ \qquad
    Alexander Prutsch$^{1}$ \qquad \\
    Wei Lin$^{3}$ \qquad
    Samuel Schulter$^{4,}$\thanks{This work is independent of the author’s employment at Amazon} \qquad
    Horst Possegger$^{1,2}$ \\ \\
    $^1$Institute of Visual Computing, Graz University of Technology \\
    $^2$Christian Doppler Laboratory for Embedded Machine Learning \\
    $^3$Institute for Machine Learning, Johannes Kepler University Linz \quad $^4$Amazon \\
    {\tt\small \{dusan.malic, reisinger, alexander.prutsch, possegger\}@tugraz.at}
}

\begin{document}

\maketitle

\begin{abstract}
This technical report outlines the top-ranking solution for RoboSense 2025: Track 3, achieving state-of-the-art performance on 3D object detection under various sensor placements.
Our submission utilizes GBlobs, a local point cloud feature descriptor specifically designed to enhance model generalization across diverse LiDAR configurations.
Current LiDAR-based 3D detectors often suffer from a \enquote{geometric shortcut} when trained on conventional global features (\ie, absolute Cartesian coordinates).
This introduces a position bias that causes models to primarily rely on absolute object position rather than distinguishing shape and appearance characteristics.
Although effective for in-domain data, this shortcut severely limits generalization when encountering different point distributions, such as those resulting from varying sensor placements.
By using GBlobs as network input features, we effectively circumvent this geometric shortcut, compelling the network to learn robust, object-centric representations.
This approach significantly enhances the model's ability to generalize, resulting in the exceptional performance demonstrated in this challenge.
\end{abstract}

\section{Introduction}
\label{sec:intro}

The majority of LiDAR-based 3D object detection architectures \cite{liu2023bevfusion,chen2023voxelnext,yin2021center,shi2023pv,shi2019pointrcnn,yan2018second} rely on global input features, specifically, the absolute Cartesian coordinates of the points.
Because of their ease of use and strong in-domain performance, global features have become the standard input representation for most leading detection models.
However, these models suffer from a geometric shortcut~\cite{wu2025sonata} exhibiting a significant bias toward object location rather than local characteristics like shape or appearance~\cite{cvpr2025gblobs}.
Consequently, this significantly limits their ability to generalize to environments with different object location distributions, such as those induced by different sensor placements.

Our contribution to the RoboSense 2025 challenge demonstrates that leveraging local point cloud geometry can substantially boost model generalization across diverse sensor configurations.
Specifically, we employ GBlobs~\cite{cvpr2025gblobs}, a novel representation that treats local neighborhoods as Gaussian blobs, defined by their mean and covariance matrices.
This formulation enables the model's encoding to be independent of the object's absolute position, effectively removing the geometric shortcut and directing the model's learning toward localized attributes, such as the shape and appearance of the objects of interest.

Calculating GBlobs requires a minimum of three points in close proximity.
Due to the inherent sparsity of LiDAR data, this requirement frequently leads to degeneracy in the far range, where local neighborhoods often lack sufficient points.
To effectively mitigate this issue, we employ a hybrid detection strategy: a secondary model is trained using conventional global Cartesian coordinates and is deployed for far-range predictions.
Both the primary (GBlobs-based) and secondary (global-coordinate-based) models are independently processed with Test-Time Augmentation (TTA).
We then revert the augmentations and apply Non-Maximum Suppression (NMS) to the output of each model separately.
Finally, the resulting predictions are spatially fused using a distance-based threshold: predictions from the GBlobs-trained model are utilized for the near-range (up to 30m), while predictions from the global-coordinate model are used exclusively beyond this threshold.

Our approach ranked 1$^\text{st}$ in the RoboSense Challenge 2025 Track 3: Sensor Placement.
This work aims to highlight the potential of local geometric features to significantly enhance model generalization, a topic we believe remains critically underexplored.
This report not only confirms this potential but also delivers a detailed explanation and analysis of the techniques employed.

\definecolor{method_red}{HTML}{A11173}
\definecolor{method_blue}{HTML}{00728F}

\begin{figure*}
\resizebox{\textwidth}{!}
{
\begin{tikzpicture}[
    nn block/.style={
        draw, 
        rectangle, 
        minimum width=2.5cm, 
        minimum height=3cm, 
        draw=method_blue,
        fill=method_blue!30, 
        font=\bfseries,
        align=center
    },
    input block/.style={
        draw, 
        rectangle, 
        minimum width=0.7cm, 
        minimum height=3cm, 
        draw=method_red,
        fill=method_red!30, 
        font=\bfseries,
        align=center
    },
    separator/.style={
        very thick,, 
        dashed, 
        gray,
    },
]
\node (Anchor) at (0,0)  {};

\node [nn block] (BEVFusionGBlobs) [above=0.3cm of Anchor]
    {\rotatebox{90}{BEVFusion-L~\cite{liu2023bevfusion}}};
\node [nn block] (BEVFusion) [below=0.3cm of Anchor]
    {\rotatebox{90}{BEVFusion-L~\cite{liu2023bevfusion}}};
    
\node [input block] (GBlobs) [left=0.2cm of BEVFusionGBlobs]
    {\rotatebox{90}{GBlobs~\cite{cvpr2025gblobs}}};

\node (AugFrame1) [left=2.6cm of Anchor]
    {\includegraphics[width=.20\textwidth,trim={300px 100px 500px 100px},clip]{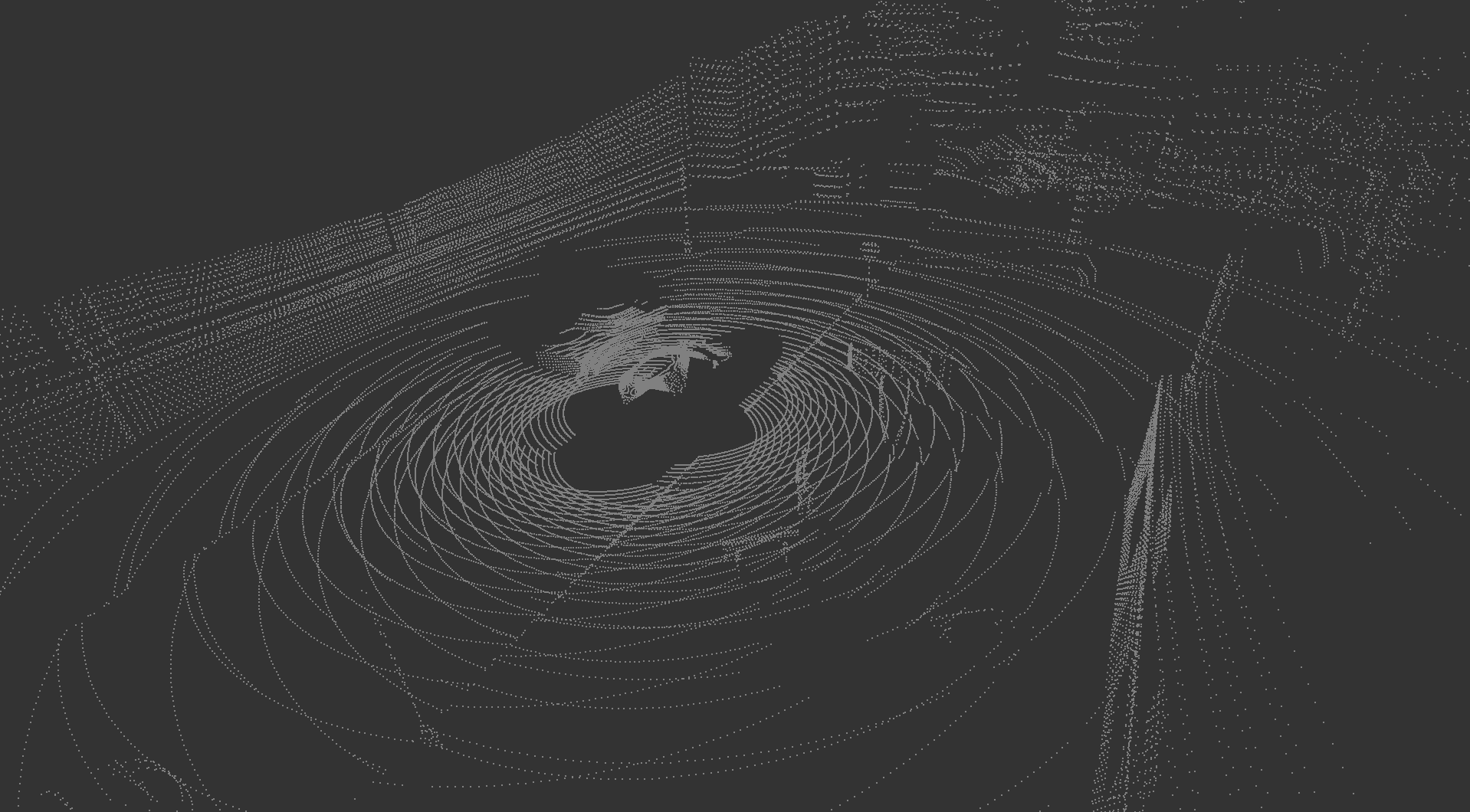}};
\node (AugFrame2) [right=-0.2cm of AugFrame1.west,  yshift=-0.2cm]
    {\includegraphics[width=.20\textwidth,trim={300px 100px 500px 100px},clip]{figures/method/input_frame.png}};
\node (AugFrameN) [right=-0.8cm of AugFrame1.west,  yshift=-0.8cm]
    {\includegraphics[width=.20\textwidth,trim={300px 100px 500px 100px},clip]{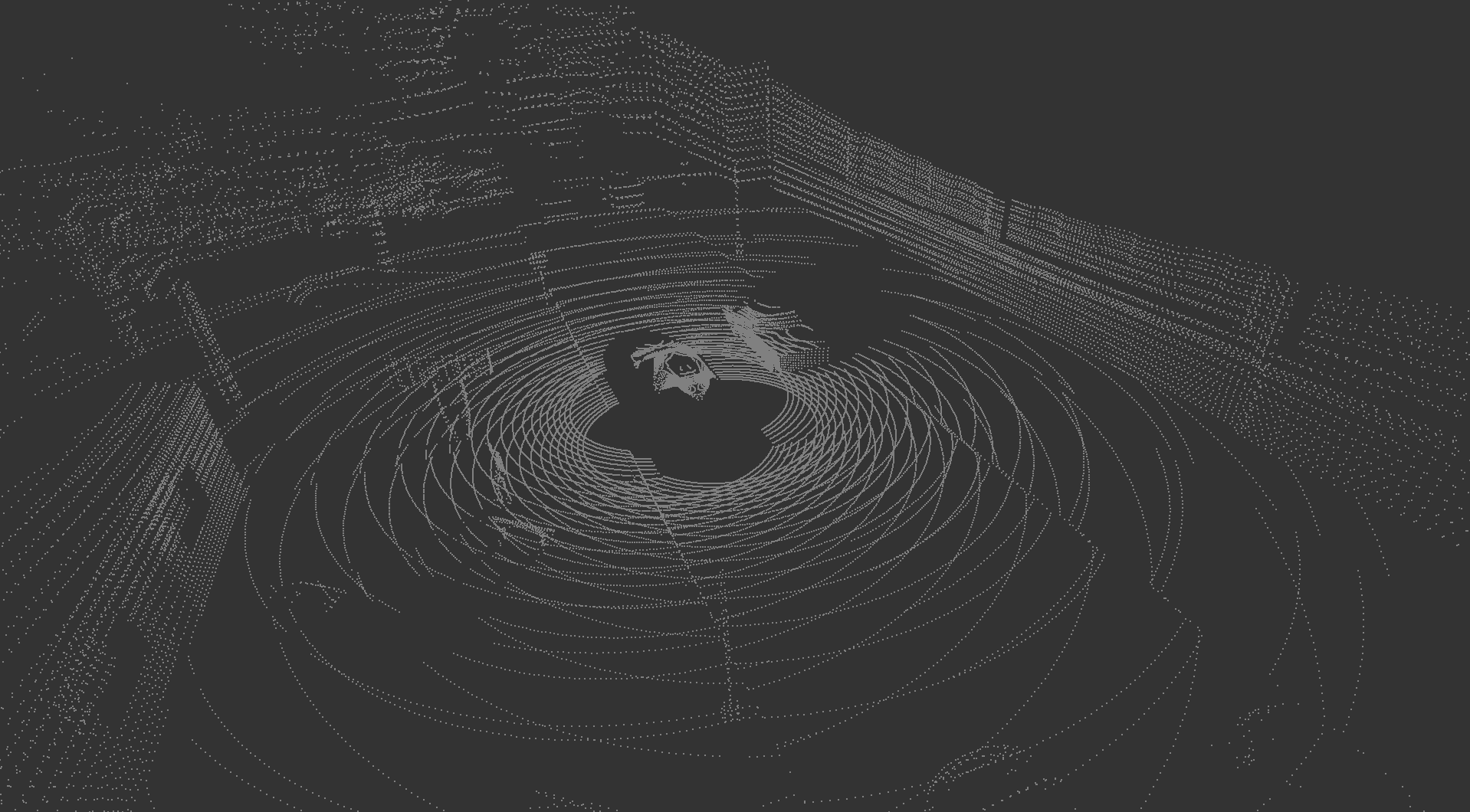}};
\node (AugDots) [above=0cm of AugFrameN.north west, xshift=0.35cm] {\rotatebox{45}{...}};

\node (InputFrame) [left=1cm of AugFrame2]
    {\includegraphics[width=.20\textwidth,trim={300px 100px 500px 100px},clip]{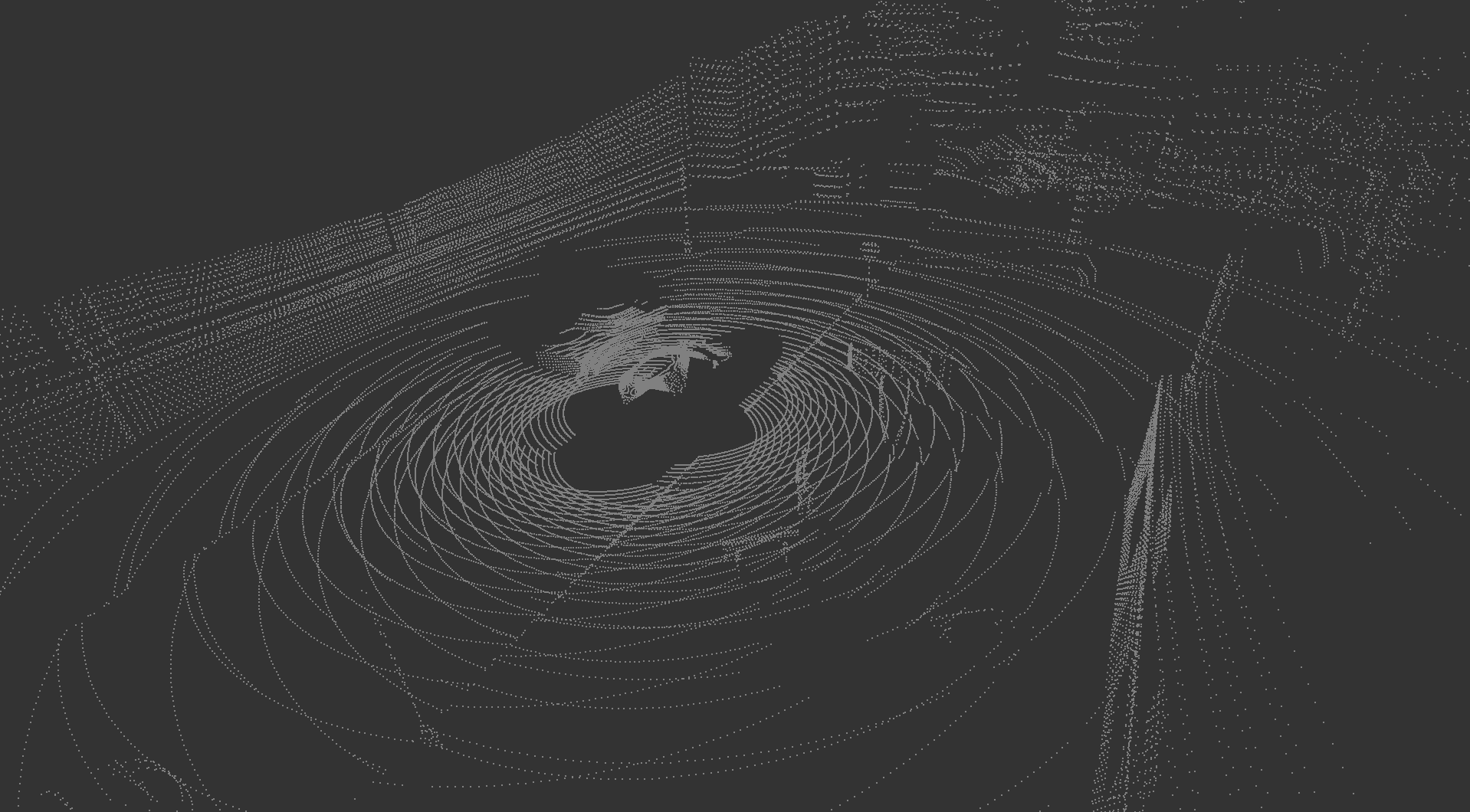}};
    
\node (OutGBlobsFrame1) [right=of BEVFusionGBlobs, yshift=0.5cm]
    {\includegraphics[width=.20\textwidth,trim={300px 100px 500px 100px},clip]{figures/method/input_frame.png}};
\node (OutGBlobsFrame2) [right=-0.2cm of OutGBlobsFrame1.west,  yshift=-0.2cm]
    {\includegraphics[width=.20\textwidth,trim={300px 100px 500px 100px},clip]{figures/method/input_frame.png}};
\node (OutGBlobsFrameN) [right=-0.8cm of OutGBlobsFrame1.west,  yshift=-0.8cm]
    {\includegraphics[width=.20\textwidth,trim={300px 100px 500px 100px},clip]{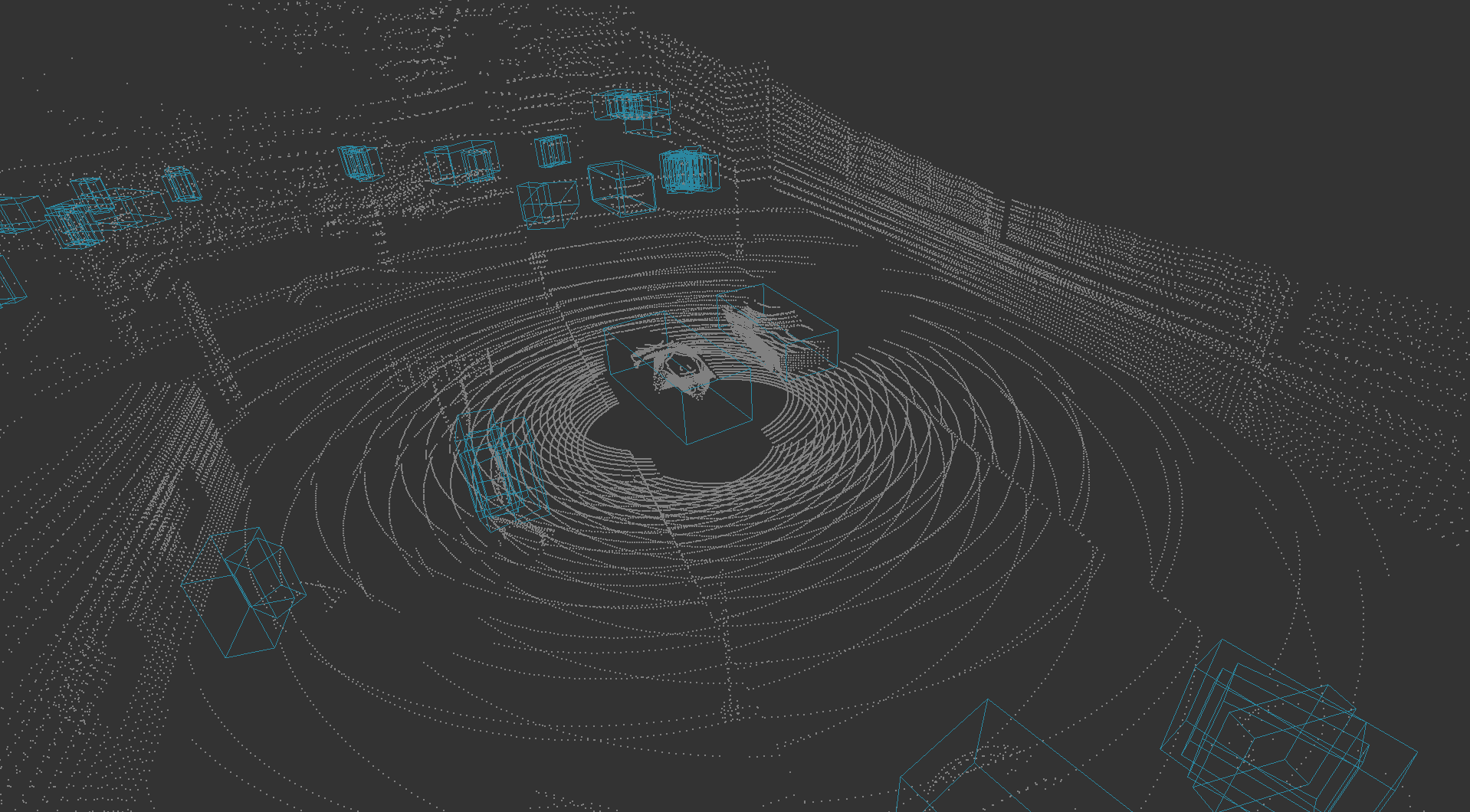}};
\node (OutDots) [above=0cm of OutGBlobsFrameN.north west, xshift=0.35cm] {\rotatebox{45}{...}};

\node (OutFrame1) [right=of BEVFusion, yshift=0.5cm]
    {\includegraphics[width=.20\textwidth,trim={300px 100px 500px 100px},clip]{figures/method/input_frame.png}};
\node (OutFrame2) [right=-0.2cm of OutFrame1.west,  yshift=-0.2cm]
    {\includegraphics[width=.20\textwidth,trim={300px 100px 500px 100px},clip]{figures/method/input_frame.png}};
\node (OutFrameN) [right=-0.8cm of OutFrame1.west,  yshift=-0.8cm]
    {\includegraphics[width=.20\textwidth,trim={300px 100px 500px 100px},clip]{figures/method/method_augmented_preds.png}};
\node (OutDots) [above=0cm of OutFrameN.north west, xshift=0.35cm] {\rotatebox{45}{...}};

\node (OutputGBlobsFrame) [right=0.5cm of OutGBlobsFrame2]
    {\includegraphics[width=.20\textwidth,trim={300px 100px 500px 100px},clip]{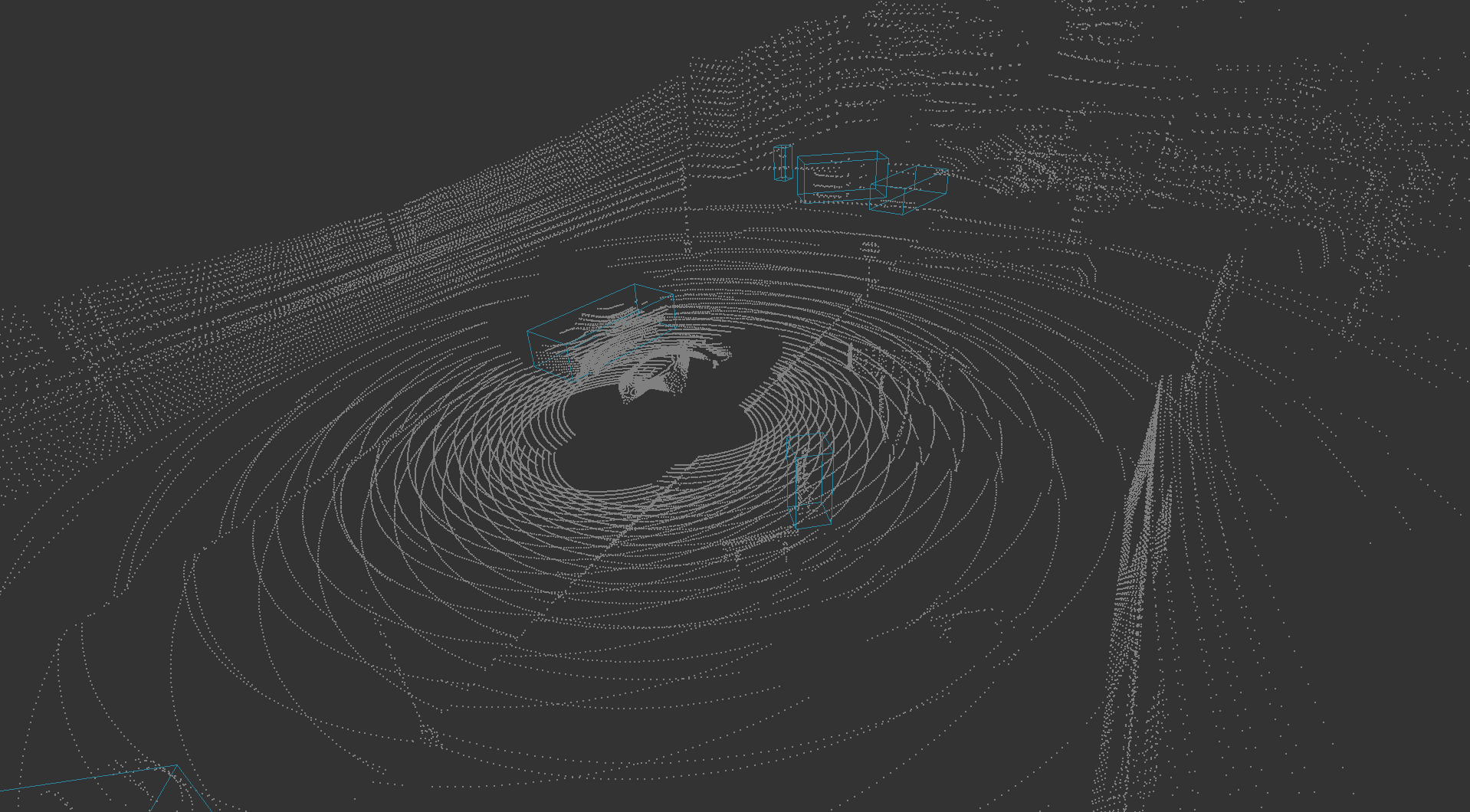}};
    
\node (OutputFrame) [right=0.5cm of OutFrame2]
    {\includegraphics[width=.20\textwidth,trim={300px 100px 500px 100px},clip]{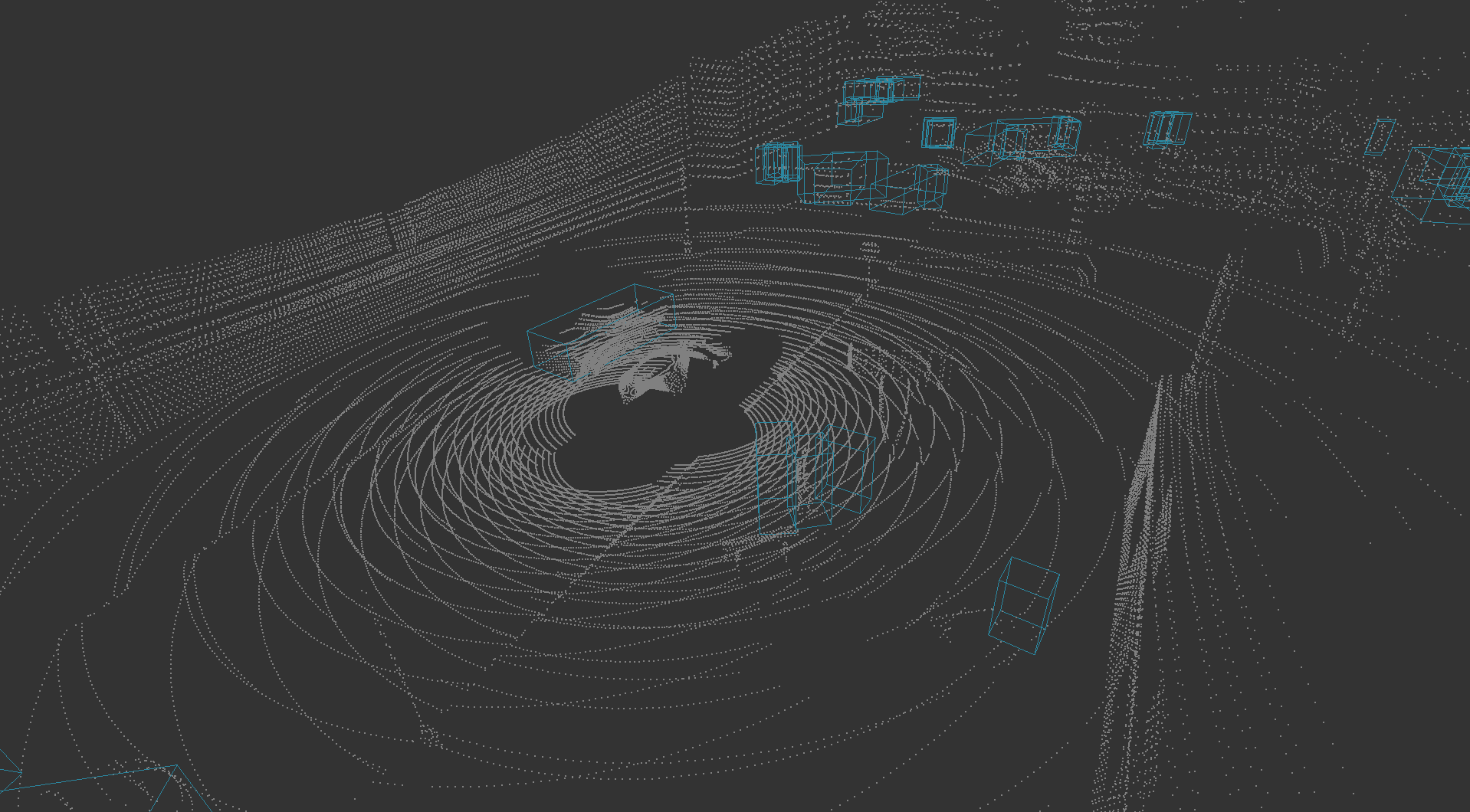}};

\node (FinalOutputFrame) [right=10.2cm of Anchor]
    {\includegraphics[width=.20\textwidth,trim={300px 100px 500px 100px},clip]{figures/method/method_final_pred.png}};
    
\draw [separator] ($(InputFrame.north east) + (0.2cm, 2.7cm)$) -- ++(0, -7.7cm);
\draw [separator] ($(AugFrame2.north east) + (0.4cm, 2.7cm)$) -- ++(0, -7.7cm);
\draw [separator] ($(AugFrame2.north east) + (9.1cm, 2.7cm)$) -- ++(0, -7.7cm);
\draw [separator] ($(AugFrame2.north east) + (13.1cm, 2.7cm)$) -- ++(0, -7.7cm);

\node(InputTitle) at ($(InputFrame.north west) + (1.3cm, 2.6cm)$) {\textcircled{1} Input point cloud};
\node(TTATitle) at ($(InputFrame.north east) + (0.9cm, 2.6cm)$) {\textcircled{2} TTA};
\node(InferenceTitle) at ($(AugFrame2.north east) + (1.5cm, 2.6cm)$) {\textcircled{3} Inference};
\node(NMSTitle) at ($(AugFrame2.north east) + (10.8cm, 2.6cm)$) {\textcircled{4} Rev. aug. \& NMS};
\node(FinalTitle) at ($(AugFrame2.north east) + (14.2cm, 2.6cm)$) {\textcircled{5} $\delta_d$ fusion    };

\end{tikzpicture}
}
\caption{
    Our method first takes an input point cloud \textcircled{1} and generates a batch of randomly augmented frames \textcircled{2}.
    This batch is then inferred \textcircled{3} by two models: the BEVFusion-L~\cite{liu2023bevfusion} baseline model and the same model trained with GBlobs~\cite{cvpr2025gblobs}.
    The augmented predictions are then reversed (de-augmented) and combined using Non-Maximum Suppression (NMS) \textcircled{4}.
    Finally, the predictions from the two models are fused \textcircled{5} via $\delta_t$ fusion, where the GBlobs-trained predictions are used up to the $\delta_t$ distance threshold, and the standard Cartesian model predictions are used beyond it.
    Best viewed on a monitor and zoomed in for detail.
}
\label{fig:method}
\end{figure*}

\section{Method}
\label{sec:method}

A significant challenge towards generalizable 3D object detection across different sensor placements is the susceptibility of deep learning models to geometric shortcuts.
When networks are trained directly on absolute Cartesian coordinates, they often learn to over-rely on an object's absolute position within the scene, rather than its inherent geometric and structural properties.
This reliance on global coordinates hinders the model's robustness and generalization to different sensor placements.

To address geometric shortcuts, we decouple the network's learning from absolute object positions.
Instead of using global Cartesian coordinates, we encode the local geometric information of the point cloud.
More precisely, given an input point cloud $X = \{\bm{p}_j = (x, y, z)\}_{j=1}^M$ of $M$ points specified by their global 3D Cartesian coordinates, we represent a local neighborhood of $N$ points as a GBlob~\cite{cvpr2025gblobs}, characterized by its mean $\bm{\mu}$ and covariance $\Sigma$, denoted as $\mathcal{N} (\bm{\mu}, \Sigma)$, where
\begin{equation}
  \bm{\mu} = \frac{1}{N} \sum_{i=1}^{N} \bm{p}_i, \quad\text{and}
  \label{eq:gauss mean}
\end{equation}
\begin{equation}
  \Sigma = \frac{1}{N} \sum_{i=1}^{N} (\bm{p}_i - \bm{\mu}) (\bm{p}_i - \bm{\mu})^\top \text{.}
  \label{eq:gauss cov}
\end{equation}
By transforming the input from absolute coordinates to GBlobs, the network is compelled to learn from the shape and local geometric structure of the object, thereby mitigating the geometric shortcut problem.

A major challenge in long-range object detection from LiDAR data is the sparse nature of point clouds.
In these sparse regions, a local neighborhood often contains only a single point, which makes it impossible to compute a meaningful covariance matrix.
As a result, our GBlobs representation degenerates into a mean-only feature, severely limiting the effectiveness of local encoding and hindering the performance of the detector.

To overcome this limitation, we introduce a dual-model approach.
We augment the GBlobs-based detector with a parallel detector that directly processes standard Cartesian coordinates.
This parallel architecture ensures robust performance even when the local encoding of the GBlobs model is compromised by point sparsity.

We enhance the robustness of both detectors using Test-Time Augmentation (TTA). We augment the input point cloud with various transformations, including translation, rotation, and scaling. Detections are then inferred from these augmented inputs and transformed back to the original coordinate system. Non-Maximum Suppression (NMS) is then applied to aggregate the predictions and remove redundant or noisy estimates.

For the final prediction, we fuse the outputs from both detectors using a simple range-based thresholding scheme.
We define a distance threshold, $\delta_d = 30$ meters, to combine the predictions: detections from the GBlobs model are used for objects within this range, while predictions from the Cartesian-coordinate model are used for all objects beyond it.
We summarize our method in \cref{fig:method}. 
This straightforward yet effective fusion strategy leverages the strengths of each model, maintaining high performance across the full range of point densities and distances.

\section{Experiments}

\begin{figure*}
\centering
\subfloat[A sample from \textit{train} split]{
    \includegraphics[width=\columnwidth]{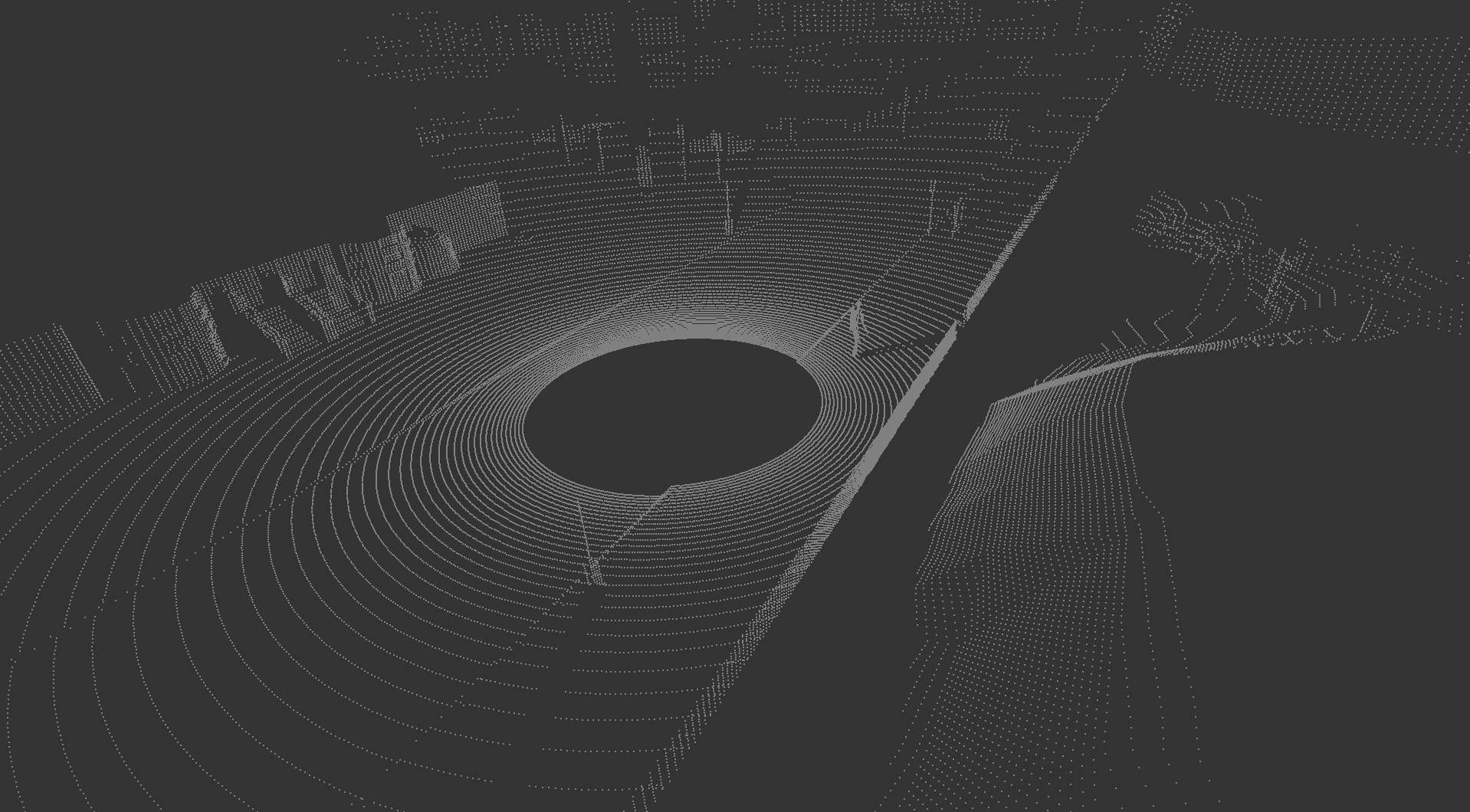}
    \label{fig:dataset:train}
}
\subfloat[A sample from \textit{test} split]{
    \includegraphics[width=\columnwidth]{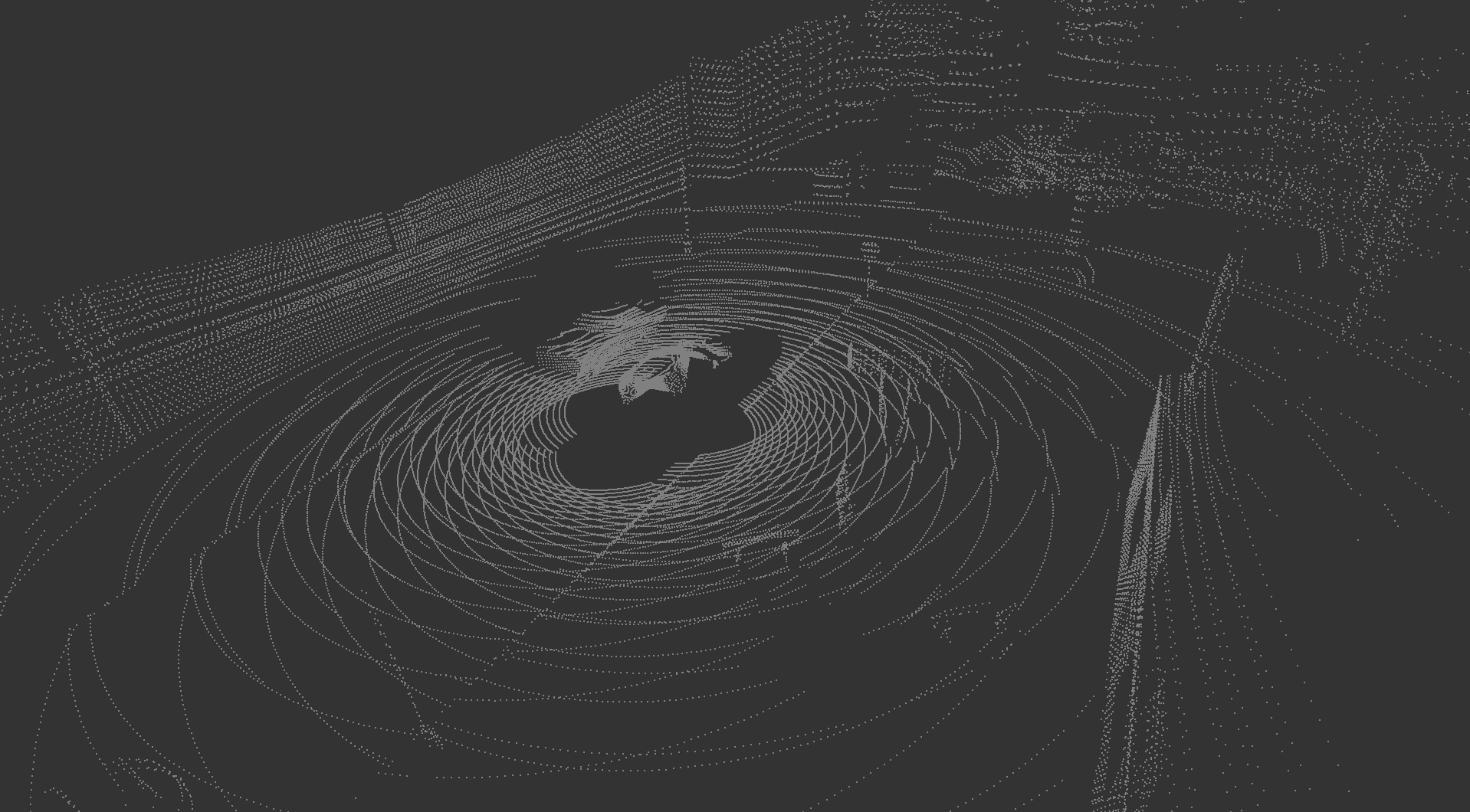}
    \label{fig:dataset:test}
}
\caption{
    Exemplary LiDAR frames from the RoboSense Challenge 2025, Track 3: Sensor Placement dataset.
}
\label{fig:dataset}
\end{figure*}

In the following section we provide the detailed experimental setup (\cref{sec:experimental_setup}), implementation details (\cref{sec:implementation_details}) and our ablation studies (\cref{sec:ablation_study}).

\subsection{Experimental Setup}
\label{sec:experimental_setup}

\paragraph{Dataset}
For the RoboSense 2025 Challenge: Track 3, the organizers provided a synthetic LiDAR dataset~\cite{li2024place3d} generated using CARLA~\cite{Dosovitskiy17}.
This dataset comprises \num{18000} samples, partitioned into a training, validation, and test split of \num{5000}, \num{3000}, and \num{10000} samples, respectively.
A key challenge of this track was overcoming the domain shift induced by different sensor placements. 
The public splits, used for training and validation, featured four known sensor placements.
In contrast, the test split introduced six completely unseen and distinct placements.
As depicted in \cref{fig:dataset}, these configuration changes translate into significant differences in the resulting point clouds.
To ensure a fair evaluation of robustness and generalization, the ground truth labels and sensor positions for the training and validation sets were released; however, the ground truth labels and the corresponding sensor positions for the test set remained confidential.

\paragraph{Metrics}
The final ranking of submissions is determined by the mean Average Precision (mAP).
To ensure robustness against statistically minor variations, a secondary metric is utilized exclusively for breaking ties.
Specifically, if the absolute difference in mAP between two submissions is within $0.01$ (one percentage point), the tie is resolved using the NuScenes Detection Score (NDS).
Thus, mAP establishes the principal rank order, with NDS serving as the decisive criterion when primary performance metrics are nearly equivalent.

\paragraph{Detector}
We follow the challenge baseline and utilize BEVFusion-L~\cite{liu2023bevfusion} as our detector.
Unlike the original BEVFusion, which fuses features from both camera and LiDAR sensors, BEVFusion-L is a variant that operates as a LiDAR-only detector.
This model leverages a spatial cross-attention mechanism to transform raw point cloud features into a bird's-eye-view (BEV) representation, which is then used for 3D object detection.

\subsection{Implementation Details}
\label{sec:implementation_details}

Our implementation is publicly available on GitHub at \href{https://github.com/malicd/GBlobs/tree/robosense25_track3}{github.com/malicd/GBlobs}\footnote{Our implementation is based on \href{https://github.com/open-mmlab/OpenPCDet}{OpenPCDet}.} for transparency and reproducibility.
We use BEVFusion-L with a combined training and validation dataset, employing a class-balanced sampling~\cite{zhu2019class} strategy to address class imbalance.
The model is trained for 90 epochs using the Adam optimizer~\cite{adam} and a cyclical learning rate scheduler with a maximum learning rate of $0.001$.
We define the LiDAR range as $[-108.0, -108.0, -5.0, 108.0, 108.0, 3.0]$ meters, using a voxel size of $[0.075, 0.075, 0.2]$ meters and accumulate 10 consecutive frames for both training and inference.
To enhance model generalization, we incorporate standard augmentation techniques, including ground truth sampling~\cite{yan2018second}, random rotation, translation, and scaling.
We disable ground truth sampling for the final 5 epochs to stabilize training.
To optimize performance for the mAP metric, we down-weight the size and rotation head losses by a factor of $0.05$ and do not predict object velocity.

For inference, we employ Test-Time Augmentation (TTA), augmenting each input frame $10$ times for both the global and Gblobs detectors with random $\pm 60^\circ$ rotation, $x$ and $y$ axis flip, and $[0.95, 1.05]$ scaling.
The augmented frames are processed to generate predictions, and Non-Maximum Suppression (NMS) is applied to consolidate detections for each individual detector.
Finally, we merge the predictions from both detectors based on a distance threshold.
We select all predictions from the Gblobs detector within $\delta_d=30$ meters and combine them with predictions from the global detector that are beyond this threshold.

\subsection{Ablation Study}
\label{sec:ablation_study}

\begin{table*}
  \centering
  \begin{tabular}{l*{9}{c}}
  \toprule
  Method & GBlobs & TTA & car & truck & bus & motorcycle & bicycle & pedestrian & mAP \\
  \midrule 
  BEVFusion-L & \ding{55} & \ding{55} & 0.9260 & 0.9013 & 0.8705 & 0.9021 & 0.8320 & 0.9020 & 0.8790 \\
  BEVFusion-L & \ding{51} & \ding{55} & 0.9244 & 0.9194 & 0.8726 & 0.9171 & 0.8400 & 0.9004 & 0.8957 \\
  BEVFusion-L & \ding{51} & \ding{51} & \bestresult{0.9329} & \bestresult{0.9252} & \bestresult{0.8758} & \bestresult{0.9199} & \bestresult{0.8416} & \bestresult{0.9460} & \bestresult{0.9069} \\
  \bottomrule
  \end{tabular}
  \caption{
    We evaluate the contribution of key components, including \emph{GBlobs}~\cite{cvpr2025gblobs} and \emph{Test-Time Augmentation (TTA)}, using the BEVFusion-L~\cite{liu2023bevfusion} baseline.
    Results are reported as per-class Average Precision (AP) and mean Average Precision (mAP) on the \emph{validation set}.
    \textbf{Bold} indicates the best performance.
    }
  \label{tab:ablation}
\end{table*}

We analyze the impact of our core design choices: utilizing GBlobs~\cite{cvpr2025gblobs} as model inputs and employing Test-Time Augmentation (TTA) to enhance prediction robustness.
All models are trained on the training split of the RoboSense 2025 Challenge: Track 3 dataset and evaluated on the corresponding validation split.
Note, however, that the dataset's test split (used for the official challenge) contains significantly different sensor placements than across the training and validation data. Consequently, the following ablation findings on the improved generalization capabilities are even more pronounced on the official test set.

The baseline model for our ablation study uses the standard (global Cartesian) input representation without TTA.
Models trained with GBlobs circumvent the geometric shortcut, which allows them to generalize better across different sensor placements. 
This benefit is clearly demonstrated in the second row of \cref{tab:ablation}.
Replacing the standard input with GBlobs, without any other modifications, yields a significant performance gain.
Specifically, the model trained with GBlobs improves the baseline performance by $1.67$ AP points.
To further boost the model's accuracy, we investigate the contribution of Test-Time Augmentation (TTA).
Applying TTA to the GBlobs-based model provides an additional improvement of $1.12$ AP points.
This confirms TTA's role in stabilizing predictions and further increasing robustness.
Overall, the synergistic effect of both design choices is substantial.
The combination of GBlobs input and TTA improves the baseline model's performance by a total of $2.79$ AP points, demonstrating the effectiveness of our proposed approach for the RoboSense 2025 challenge.

As detailed in \cref{sec:method}, in sparse LiDAR regions (\eg, far-range), a local neighborhood often contains only a single point.
This makes it impossible to compute a covariance matrix as per \cref{eq:gauss cov}.
Consequently, our GBlobs representation degenerates into a mean-only feature, severely limiting the effectiveness of local encoding and hindering the performance of the detector.
To mitigate this, we employ a hybrid approach: we utilize GBlobs predictions up to a distance threshold $\delta_d$ and use predictions from the standard global-feature baseline model beyond this threshold.
To determine an optimal $\delta_d$, we performed an ablation study by evaluating the performance while gradually removing objects up to a certain distance.
In \cref{fig:distance_evaluation}, we observe that the GBlobs-based model outperforms the baseline up to approximately $28$ meters.
However, as sparsity increases with distance, the GBlobs model's performance then rapidly deteriorates.
Based on this trade-off, we select $\delta_d$ to be $30$ meters for the final evaluation, using GBlobs-based detections within this range and baseline predictions for all objects beyond it.

\definecolor{matplotlib_blue}{HTML}{1f77b4}
\definecolor{matplotlib_orange}{HTML}{ff7f0e}

\definecolor{method_red}{HTML}{A11173}
\definecolor{method_blue}{HTML}{00728F}

\begin{figure}
\begin{tikzpicture}
\begin{axis}[
    width=\columnwidth,
    xlabel={Distance [m]},
    ylabel={mAP},
    xmin=5, xmax=45,
    ymin=69, ymax=86,
    legend pos=south west,
    ymajorgrids=true,
    grid style=dashed,
    legend cell align={left}, %
    legend entries={Global, GBlobs~\cite{cvpr2025gblobs}},
]

\addplot[
    color=method_blue,
    mark=o,
    line width=1.2pt,
    ]
    coordinates {
   (10,84.16)(20,81.20)(30,76.71)(40,69.88)
    };\label{fig:tikz:global}

\addplot[
    color=method_red,
    mark=square,
    line width=1.2pt,
    ]
    coordinates {
   (10,84.72)(20,81.66)(30,76.35)(40,69.29)
    };\label{fig:tikz:gblobs}
    
\end{axis}
\end{tikzpicture}
\caption{
    We evaluate two BEVFusion-L~\cite{liu2023bevfusion} models: a standard model trained with global input features (\ref{fig:tikz:global}) and a model trained with GBlobs~\cite{cvpr2025gblobs} (\ref{fig:tikz:gblobs}).
    The $x$-axis represents the distance cut-off, where predictions and ground truth below this value are ignored during evaluation.
    The $y$-axis reports the mean Average Precision (mAP) computed across all detection classes on the validation split.
    Both models were trained on the training split of the RoboSense 2025 Challenge: Track 3 dataset.
}
\label{fig:distance_evaluation}
\end{figure}
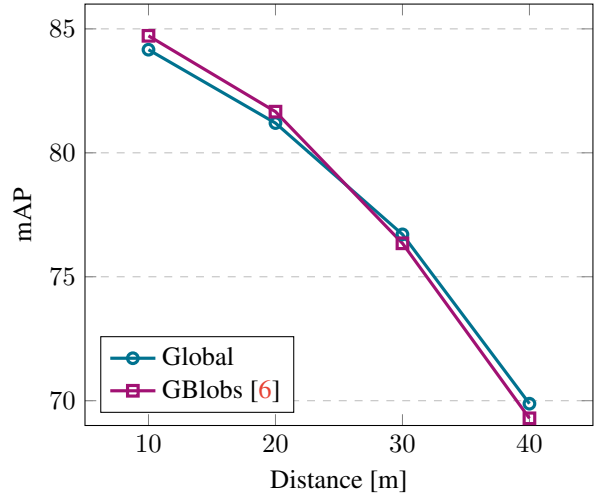

\section{Conclusion}

Our solution for the RoboSense 2025: Track 3 competition showcases a robust approach to 3D object detection with varied sensor placements.
By leveraging GBlobs, we demonstrate a generalizable model that maintains high robustness across diverse sensor configurations.
This work not only provides a high-performing solution but also identifies a key direction for the field: improving model generalization with local features, especially in sparse areas.
We hope our methodology will serve as a strong foundation for future research in this domain.

\section*{Acknowledgments}
We gratefully acknowledge the financial support by the Austrian Federal Ministry for Digital and Economic Affairs, the National Foundation for Research, Technology and Development and the Christian Doppler Research Association.
We further acknowledge the EuroHPC Joint Undertaking for awarding us access to Leonardo at CINECA, Italy.

{
\small
\bibliographystyle{ieeenat_fullname}
\bibliography{abbrv_short,main}

\begin{thebibliography}{12}
\providecommand{\natexlab}[1]{#1}
\providecommand{\url}[1]{\texttt{#1}}
\expandafter\ifx\csname urlstyle\endcsname\relax
  \providecommand{\doi}[1]{doi: #1}\else
  \providecommand{\doi}{doi: \begingroup \urlstyle{rm}\Url}\fi

\bibitem[Chen et~al.(2023)Chen, Liu, Zhang, Qi, and Jia]{chen2023voxelnext}
Yukang Chen, Jianhui Liu, Xiangyu Zhang, Xiaojuan Qi, and Jiaya Jia.
\newblock {VoxelNeXt: Fully Sparse VoxelNet for 3D Object Detection and Tracking}.
\newblock In \emph{Proc. CVPR}, 2023.

\bibitem[Dosovitskiy et~al.(2017)Dosovitskiy, Ros, Codevilla, Lopez, and Koltun]{Dosovitskiy17}
Alexey Dosovitskiy, German Ros, Felipe Codevilla, Antonio Lopez, and Vladlen Koltun.
\newblock {{CARLA}}: {{An Open Urban Driving Simulator}}.
\newblock In \emph{Proc. {{CoRL}}}, 2017.

\bibitem[Kingma and Ba(2015)]{adam}
Diederik~P. Kingma and Jimmy Ba.
\newblock {Adam: A Method for Stochastic Optimization}.
\newblock In \emph{Proc. ICLR}, 2015.

\bibitem[Li et~al.(2024)Li, Kong, Hu, Xu, and Huang]{li2024place3d}
Ye Li, Lingdong Kong, Hanjiang Hu, Xiaohao Xu, and Xiaonan Huang.
\newblock Is your lidar placement optimized for 3d scene understanding?
\newblock In \emph{Proc. NeurIPS}, 2024.

\bibitem[Liu et~al.(2023)Liu, Tang, Amini, Yang, Mao, Rus, and Han]{liu2023bevfusion}
Zhijian Liu, Haotian Tang, Alexander Amini, Xinyu Yang, Huizi Mao, Daniela~L Rus, and Song Han.
\newblock {BEVFusion: Multi-Task Multi-Sensor Fusion with Unified Bird's-Eye View Representation}.
\newblock In \emph{Proc. ICRA}, 2023.

\bibitem[Mali\'c et~al.(2025)Mali\'c, Fruhwirth-Reisinger, Schulter, and Possegger]{cvpr2025gblobs}
Du\v{s}an Mali\'c, Christian Fruhwirth-Reisinger, Samuel Schulter, and Horst Possegger.
\newblock {GBlobs: Explicit Local Structure via Gaussian Blobs for Improved Cross-Domain LiDAR-based 3D Object Detection}.
\newblock In \emph{Proc. CVPR}, 2025.

\bibitem[Shi et~al.(2019)Shi, Wang, and Li]{shi2019pointrcnn}
Shaoshuai Shi, Xiaogang Wang, and Hongsheng Li.
\newblock {PointRCNN: 3D Object Proposal Generation and Detection from Point Cloud}.
\newblock In \emph{Proc. CVPR}, 2019.

\bibitem[Shi et~al.(2023)Shi, Jiang, Deng, Wang, Guo, Shi, Wang, and Li]{shi2023pv}
Shaoshuai Shi, Li Jiang, Jiajun Deng, Zhe Wang, Chaoxu Guo, Jianping Shi, Xiaogang Wang, and Hongsheng Li.
\newblock {PV-RCNN++: Point-Voxel Feature Set Abstraction With Local Vector Representation for 3D Object Detection}.
\newblock \emph{IJCV}, 131\penalty0 (2):\penalty0 531--551, 2023.

\bibitem[Wu et~al.(2025)Wu, DeTone, Frost, Shen, Xie, Yang, Engel, Newcombe, Zhao, and Straub]{wu2025sonata}
Xiaoyang Wu, Daniel DeTone, Duncan Frost, Tianwei Shen, Chris Xie, Nan Yang, Jakob Engel, Richard Newcombe, Hengshuang Zhao, and Julian Straub.
\newblock {Sonata: Self-Supervised Learning of Reliable Point Representations}.
\newblock In \emph{Proc. CVPR}, 2025.

\bibitem[Yan et~al.(2018)Yan, Mao, and Li]{yan2018second}
Yan Yan, Yuxing Mao, and Bo Li.
\newblock {SECOND: Sparsely Embedded Convolutional Detection}.
\newblock \emph{Sensors}, 18\penalty0 (10):\penalty0 3337, 2018.

\bibitem[Yin et~al.(2021)Yin, Zhou, and Krahenbuhl]{yin2021center}
Tianwei Yin, Xingyi Zhou, and Philipp Krahenbuhl.
\newblock {Center-based 3D Object Detection and Tracking}.
\newblock In \emph{Proc. CVPR}, 2021.

\bibitem[Zhu et~al.(2019)Zhu, Jiang, Zhou, Li, and Yu]{zhu2019class}
Benjin Zhu, Zhengkai Jiang, Xiangxin Zhou, Zeming Li, and Gang Yu.
\newblock {Class-balanced Grouping and Sampling for Point Cloud 3D Object Detection}.
\newblock \emph{arXiv CoRR}, abs/1908.09492, 2019.

\end{thebibliography}
}

\end{document}